\documentclass{article}

\usepackage{arxiv}

\usepackage[utf8]{inputenc} 
\usepackage[T1]{fontenc}    
\usepackage{hyperref}       
\usepackage{url}            
\usepackage{booktabs}       
\usepackage{amsfonts}       
\usepackage{nicefrac}       
\usepackage{microtype}      
\usepackage{lipsum}

\usepackage{bm}
\usepackage{amsmath}
\usepackage{amssymb}
\usepackage{amsfonts}
\usepackage{graphicx}
\def\vec#1{\mathbf{#1}}
\usepackage{algorithm}
\usepackage{algorithmic}
\usepackage{comment}

\title{Meta-learning for Matrix Factorization without Shared Rows or Columns}

\author{
  Tomoharu Iwata\\
  NTT Communication Science Laboratories\\
}
\date{}

\begin{document}
\maketitle

\begin{abstract}
We propose a method that meta-learns a knowledge on matrix factorization from various matrices, and uses the knowledge for factorizing unseen matrices. The proposed method uses a neural network that takes a matrix as input, and generates prior distributions of factorized matrices of the given matrix. The neural network is meta-learned such that the expected imputation error is minimized  when the factorized matrices are adapted to each matrix by a maximum a posteriori (MAP) estimation. We use a gradient descent method for the MAP estimation, which enables us to backpropagate the expected imputation error through the gradient descent steps for updating neural network parameters since each gradient descent step is written in a closed form and is differentiable. The proposed method can meta-learn from matrices even when their rows and columns are not shared, and their sizes are different from each other. In our experiments with three user-item rating datasets, we demonstrate that our proposed method can impute the missing values from a limited number of observations in unseen matrices after being trained with different matrices.
\end{abstract}

\section{Introduction}

Matrix factorization is an important machine learning technique
for imputing missing values and analyzing hidden structures
in matrices.
With matrix factorization, a matrix is modeled by the product of two low-rank matrices, assuming that the rank of the given matrix is low.
Matrix factorization has been used in
a wide variety of applications, such as collaborative filtering~\cite{bokde2015matrix,mnih2008probabilistic,salakhutdinov2008bayesian,koren2009matrix},
text analysis~\cite{dumais2004latent,hofmann2001unsupervised},
bioinformatics~\cite{brunet2004metagenes},
and spatio-temporal data analysis~\cite{kimura2014spatio,takeuchi2017structurally}.
However,
when the number of observations is not large enough,
existing matrix factorization methods fail to
impute the missing values.
In some applications, only a limited number of observations are available.
For example, a newly launched recommender system
only has histories for small numbers of users and items,
and spatio-temporal data are not accumulated in the beginning when a new region is analyzed.

Recently,
few-shot learning and meta-learning have attracted attention for
learning from few labeled data~\cite{schmidhuber:1987:srl,bengio1991learning,finn2017model,vinyals2016matching,snell2017prototypical}.
Meta-learning methods learn how to learn from a small amount of
labeled data in various tasks,
and use the learned knowledge in unseen tasks.
Existing meta-learning methods assume that attributes are the same across all tasks.
Therefore, they are inapplicable to matrix factorization
when the rows or columns are not shared across matrices,
or the matrix sizes are different across matrices.

In this paper, we propose a meta-learning method for matrix factorization,
which can learn from various matrices without shared rows or columns,
and use the learned knowledge for the missing value imputation of unseen matrices.
The meta-training and meta-test matrices contain the missing values,
and their sizes can be different from each other.
With the proposed method, the prior distributions
of two factorized matrices are modeled by a neural network
that takes a matrix as input.
We use exchangeable matrix layers~\cite{hartford2018deep}
and permutation invariant networks~\cite{zaheer2017deep} for the neural network,
with which we encode the information of the given matrix
into the priors.
For each matrix, its factorized matrices are adapted to the given matrix by
maximum a posteriori (MAP) estimation using the gradient descent method.
The posteriors are calculated using the neural network-based priors
and the observations on the given matrix based on Bayes' theorem.
Since the neural network is shared across all matrices,
we can learn shared hidden structure in various meta-training matrices,
and use it for unseen meta-test matrices.

\begin{figure}[t!]
  \centering
  \includegraphics[width=26em]{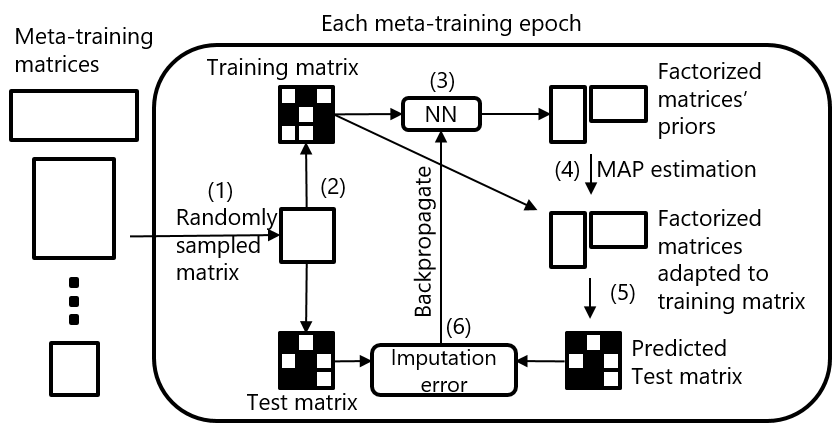}
  \caption{Our meta-learning framework: We are given multiple meta-training matrices. For each meta-training epoch, first, we generate a matrix from randomly selected rows and columns of a randomly selected matrix from the meta-training matrices. Second, we split elements in the matrix into training and test matrices. Third, factorized matrices' priors are inferred by a neural network. Fourth, we adapt the factorized matrices to the training matrix by maximizing the posterior with a gradient descent method. Fifth, we impute the missing values by multiplying the adapted factorized matrices. Sixth, we calculate the missing value imputation error using the test matrix and backpropagate it to update neural network's parameters.}
  \label{fig:framework} 
\end{figure}

We meta-learn the neural networks such that
the missing value imputation error is minimized
when the MAP estimated factorized matrices are used for imputation.
Since the gradient descent steps for the MAP estimation are differentiable,
we can backpropagate the missing value imputation error
through the MAP estimation for updating the neural network parameters in the priors.
For each meta-training epoch based on the episodic training framework~\cite{finn2017model},
training and test matrices are randomly generated from the meta-training matrices,
and the test matrix imputation error of
the factorized matrices adapted to the training matrix
is evaluated and backpropagated.
Figure~\ref{fig:framework} shows the meta-learning framework of our proposed method.
Although we explain the proposed method with matrix imputation,
it is straightforwardly extended for tensor imputation
using exchangeable tensor layers~\cite{hartford2018deep}
and tensor factorization~\cite{kuleshov2015tensor}
instead of exchangeable matrix layers and matrix factorization.

The following are our major contributions:
\begin{enumerate}
\item We propose a meta-learning method for matrix imputation that can meta-learn from matrices without shared rows or columns.
\item We design a neural network to generate prior distributions of factorized matrices with different sizes, which is meta-trained such that the test matrix imputation performance improves when the factorized matrices are adapted to the training matrix based on the MAP estimation.
\item In our experiments using real-world data sets, we demonstrate that the proposed method achieves better matrix imputation performance when meta-training data contain matrices that are related to meta-test matrices.
  \end{enumerate}

\section{Related work}
\label{sec:related}

Many meta-learning or few-shot learning methods have been proposed~\cite{schmidhuber:1987:srl,bengio1991learning,ravi2016optimization,andrychowicz2016learning,vinyals2016matching,snell2017prototypical,bartunov2018few,finn2017model,li2017meta,kimbayesian,finn2018probabilistic,rusu2018meta,yao2019hierarchically,edwards2016towards,garnelo2018conditional,kim2019attentive,hewitt2018variational,bornschein2017variational,reed2017few,rezende2016one}.
These existing methods cannot learn from matrices without shared or and columns.
With probabilistic meta-learning methods~\cite{finn2018probabilistic,kimbayesian},
the prior of model parameters is meta-trained,
where they require that the numbers of parameters are the same across tasks.
Therefore, they are inapplicable for meta-learning nonparametric models,
including matrix factorization,
where the number of parameters can grow with the sample size.
In contrast, the proposed method meta-trains a neural network
that generates the prior of a task-specific model,
which enables us to meta-learn nonparametric models.
The proposed method is related
to model-agnostic meta-learning~\cite{finn2017model} (MAML)
in the sense that both methods backpropagate a loss
through gradient descent steps.
MAML learns the initial values of model parameters
such that the performance improves when all the parameters are adapted to new tasks.
The proposed method is more efficient than MAML
since MAML requires a second-order gradient computation on the whole neural network
while the proposed method requires that only on factorized matrices.
In our model, we explicitly incorporate matrix factorization procedures
by the gradient descent method,
which has been successfully used for a wide variety of matrix imputation applications.

The proposed method is also related to encoder-decoder based
meta-learning methods~\cite{xu2020metafun},
such as neural processes~\cite{garnelo2018conditional}.
The encoder-decoder based meta-learning methods
obtain a representation of few observations by an encoder
and use it in predictions by a decoder,
where the encoder and decoder are modeled by neural networks.
Similarly, the proposed method uses a neural network to
obtain factorized matrices for predicting the missing values.
Their differences are that
the proposed method uses a neural network designed for matrices with missing values, and
has gradient descent steps for adapting factorized matrices to observations.
Adapting parameters by directly fitting them to observations
is effective for meta-learning~\cite{lee2019meta,iwata2020few}
since it is difficult to output the adapted parameters
only by neural networks for a wide variety of observations.
The proposed method uses exchangeable matrix layers~\cite{hartford2018deep}
as its components.
The exchangeable matrix layers have not been used for meta-learning.
Heterogeneous meta-learning~\cite{iwata2020meta} can learn from multiple tasks
without shared attributes. However, it cannot handle missing values,
and is inapplicable to matrix imputation.

Collective matrix factorization~\cite{singh2008relational,bouchard2013convex,yang2015robust}
simultaneously factorizes multiple matrices,
where information
in a matrix can be transferred to other matrices for factorization.
Collective matrix factorization assumes that
some of the columns and/or rows are shared across matrices.
Transfer learning methods for matrix factorization that do not assume shared columns or rows
have been proposed~\cite{iwata2015cross}.
With such transfer learning methods,
target matrices are required in the training phase
for transferring knowledge from source to target matrices.
On the other hand,
the proposed method does not need to use target matrices for training.
Several few-shot learning methods for recommender systems
have been proposed~\cite{vartak2017meta,li2020few} to
tackle the cold start problem,
where the histories of new users or new items are insufficiently accumulated.
These methods use auxiliary information,
such as user attributes and item descriptions.
In contrast, the proposed method does not use auxiliary information.

\section{Proposed method}
\label{sec:proposed}

\subsection{Problem formulation}

Suppose that we are given $D$ matrices
$\mathcal{X}=\{\vec{X}_{d}\}_{d=1}^{D}$ in the meta-training phase,
where $\vec{X}_{d}\in\mathbb{R}^{N_{d}\times M_{d}}$
is the meta-training matrix in the $d$th task,
and $x_{dnm}$ is its $(n,m)$th element.
The sizes of the meta-training matrices can be different across tasks,
$N_{d}\neq N_{d'}$, $M_{d}\neq M_{d'}$,
and rows or columns are not shared across matrices.
The meta-training matrices can contain missing values.
In this case, we are additionally given binary indicator matrix
$\vec{B}_{d}\in\{0,1\}^{N_{d}\times M_{d}}$,
where $b_{dnm}=1$ if the $(n,m)$th element is observed,
and $b_{dnm}=0$ otherwise.
In the meta-test phase,
we are given a meta-test matrix with missing values,
$\vec{X}_{*}\in\mathbb{R}^{N_{*}\times M_{*}}$,
where the missing values are specified by indicator matrix
$\vec{B}_{*}\in\{0,1\}^{N_{*}\times M_{*}}$.
Our aim is to improve the missing value imputation performance on the meta-test matrix.

\subsection{Model}
\label{sec:model}

\begin{figure*}[t!]
  \centering
  \includegraphics[width=35em]{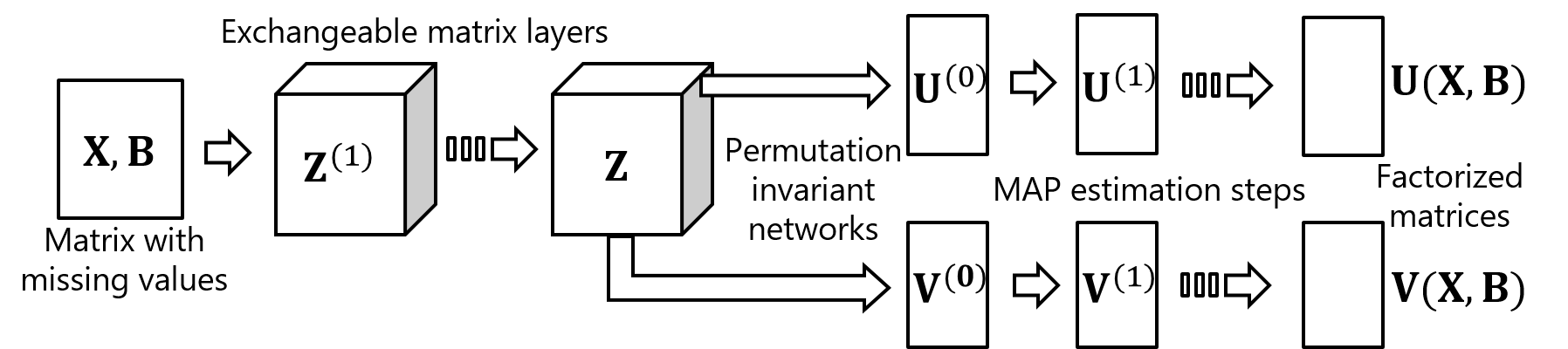}
  \caption{Our model: Matrix with missing values $\vec{X}$ and its indicator matrix $\vec{B}$ are the input. First, representations $\vec{Z}$ for each element in $\vec{X}$ are obtained by exchangeable matrix layers in Eq.~(\ref{eq:z}). Second, the mean of the prior of factorized matrices $\vec{U}^{(0)}$ and $\vec{V}^{(0)}$ is calculated by transforming the averages of representations $\vec{Z}$ over columns and rows by permutation invariant neural networks in Eq.~(\ref{eq:uv0}).
    Third, factorized matrices $\vec{U}(\vec{X},\vec{B})$ and $\vec{V}(\vec{X},\vec{B})$ are estimated by maximizing the posterior probability using the gradient descent steps in Eqs.~(\ref{eq:u},\ref{eq:v}). For each gradient descent step, input $\vec{X}$ and $\vec{B}$ and prior means $\vec{U}^{(0)}$ and $\vec{V}^{(0)}$ are used as well as previous estimates $\vec{U}^{(t-1)}$ and $\vec{V}^{(t-1)}$. Our model can take matrices with different sizes as input, and output their factorized matrices.}
  \label{fig:model} 
\end{figure*}

For each task, our model outputs
factorized matrices $\vec{U}(\vec{X},\vec{B})\in\mathbb{R}^{N\times K}$
and $\vec{V}(\vec{X},\vec{B})\in\mathbb{R}^{M\times K}$
given matrix with missing values $\vec{X}\in\mathbb{R}^{N\times M}$
and its indicator matrix $\vec{B}\in\{0,1\}^{N\times M}$.
In the meta-training phase,
$\vec{X}$ and $\vec{B}$ are generated from one of
meta-training matrices $\{\vec{X}_{d}\}_{d=1}^{D}$
as described in Section~\ref{sec:training}.
In the meta-test phase,
$\vec{X}$ and $\vec{B}$ correspond to
meta-test matrix $\vec{X}_{*}$ and its indicator matrix $\vec{B}_{*}$.
Figure~\ref{fig:model} illustrates our model.

With our model, the factorized matrices are estimated
by maximizing the posterior probability:
\begin{align}
  \vec{U}(\vec{X},\vec{B}),\vec{V}(\vec{X},\vec{B})
  = \arg\max_{\vec{U},\vec{V}} \left[\log p(\vec{X}|\vec{U},\vec{V})
  + \log p_{\vec{X}}(\vec{U},\vec{V}|\mathcal{X})\right],
  \label{eq:map}
\end{align}
where the first term is the likelihood,
\begin{align}
  \log p(\vec{X}|\vec{U},\vec{V})\propto -\sum_{n,m=1}^{N,M}b_{nm}(\vec{u}_{n}^{\top}\vec{v}_{m}-x_{nm})^{2},
\end{align}
the second term is the prior,
$\vec{U}=[\vec{u}_{1},\cdots,\vec{u}_{N}]^{\top}$,
$\vec{V}=[\vec{v}_{1},\cdots,\vec{v}_{M}]^{\top}$,
and $\top$ is the transpose.
We use subscript $\vec{X}$ in prior $p_{\vec{X}}$ to indicate that it is task-specific.
The prior is conditioned on meta-training matrices $\mathcal{X}$
since we meta-learn it from $\mathcal{X}$.
We assume a Gaussian prior with mean $\vec{u}_{n}^{(0)}$ ($\vec{v}_{m}^{(0)}$)
and variance $\lambda^{-1}$:
\begin{align}
\log p_{\vec{X}}(\vec{U},\vec{V}|\mathcal{X})\propto  
-\lambda\left(
  \sum_{n=1}^{N}\parallel\vec{u}_{n}-\vec{u}^{(0)}_{n}\parallel^{2}
  +\sum_{m=1}^{M}\parallel\vec{v}_{m}-\vec{v}^{(0)}_{m}\parallel^{2}
  \right).
\end{align}
We generate the task-specific prior mean values $\vec{u}_{n}^{(0)}, \vec{v}_{m}^{(0)}$
with different sizes using a neural network.
The neural network is shared across different tasks,
which enables us to extract knowledge in the meta-training matrices,
and use the knowledge for unseen tasks.
When $\lambda=0$, matrix factorization
is independently performed for each matrix,
and it cannot meta-learn useful knowledge for factorization.

For modeling the prior, first,
we obtain representations $\vec{Z}\in\mathbb{R}^{N\times M\times C}$
of given matrix $\vec{X}$ using a neural network,
where the matrix's $(n,m)$th element is represented
by vector $\vec{z}_{nm}\in\mathbb{R}^{C}$.
Our model uses exchangeable matrix layers~\cite{hartford2018deep}:
\begin{align}
  z^{(\ell+1)}_{nmc}&=\sigma\Biggl(\sum_{c'=1}^{C^{(\ell)}}
    \Bigl(w_{c'c1}^{(\ell)}b_{nm}z_{nmc'}^{(\ell)}
  +w_{c'c2}^{(\ell)}\frac{\sum_{n'=1}^{N} b_{n'm}z_{n'mc'}^{(\ell)}}{\sum_{n'=1}^{N} b_{n'm}}
  +w_{c'c3}^{(\ell)}\frac{\sum_{m'=1}^{M} b_{nm'}z_{nm'c'}^{(\ell)}}{\sum_{m'=1}^{M} b_{nm'}}
  \nonumber\\
  &+w_{c'c4}^{(\ell)}\frac{\sum_{n',m'=1}^{N,M} b_{n'm'}z_{n'm'c'}^{(\ell)}}{\sum_{n',m'=1}^{N,M} b_{n'm'}}
  +w_{c5}^{(\ell)}\Bigr)\Biggr),
  \label{eq:z}
\end{align}
where
$z^{(\ell)}_{nmc}\in\mathbb{R}$ is the $c$th channel of the representation
of the $(n,m)$ element in the $\ell$th layer,
$w_{c'ci}^{(\ell)}\in\mathbb{R}$ is a weight parameter in the $\ell$th layer
to be trained
for the influence of channel $c'$ on channel $c$ in the next layer,
$\sigma$ is an activation function,
and $C^{(\ell)}$ is the channel size of the $\ell$th layer.
In the first layer, the given matrix is used for representation
$\vec{z}^{(0)}_{nm}=x_{nm}\in\mathbb{R}$, where
$x_{nm}$ is the value of the $(n,m)$ element of given matrix $\vec{X}$.
The representation in the last layer
is used as final representation $\vec{Z}=\vec{Z}^{(L)}$,
where $L$ is the number of layers,
and $C^{(L)}=C$.
In the last layer, activation function $\sigma$ is omitted.
The first term in Eq.~(\ref{eq:z}) calculates the influence
from the same element,
the second term calculates the influence from the elements of the same column,
the third term calculates the influence from the elements of the same row,
the fourth term calculates the influence from all the elements,
and the fifth term is the bias.
The influences are averaged over the observed elements using indicator matrix $\vec{B}$.
The exchangeable matrix layer is permutation
equivariant, where the output is the same values
across all the row- and column-wise permutations of the input.
With the exchangeable matrix layers,
we can obtain representations for each element
considering the whole matrix.
The exchangeable matrix layers can handle matrices with different sizes
since their parameters $w^{(\ell)}_{c'ci}$ do not depend on the input matrix size.

Second, we calculate the mean of the priors of the factorized matrices
using permutation invariant networks~\cite{zaheer2017deep}:
\begin{align}
  \vec{u}^{(0)}_{n}=f_{\mathrm{U}}\left(\frac{1}{M}\sum_{m=1}^{M}\vec{z}_{nm}\right),
  \quad
  \vec{v}^{(0)}_{m}=f_{\mathrm{V}}\left(\frac{1}{N}\sum_{n=1}^{N}\vec{z}_{nm}\right),
  \label{eq:uv0}  
\end{align}
where $\vec{u}^{(0)}_{n}\in\mathbb{R}^{K}$ is the prior mean
of the $n$th row of factorized matrix $\vec{U}$,
$\vec{v}^{(0)}_{m}\in\mathbb{R}^{K}$ is the pror mean
of the $m$th row of factorized matrix $\vec{V}$,
and $f_{\mathrm{U}}$ and $f_{\mathrm{V}}$ are feed-forward neural networks.
In Eq.~(\ref{eq:uv0}), 
we take the average of element representation $\vec{Z}$
over the rows (columns) and input them into the neural networks.
Eq.~(\ref{eq:uv0}) is a permutation invariant operation
that can take any number of elements $N$ and $M$.
By Eqs.~(\ref{eq:z},\ref{eq:uv0}),
we can obtain prior mean values $\vec{u}^{(0)}_{n}$ and $\vec{v}^{(0)}_{m}$
for each row and column considering the relationship with
other elements in the given matrix.
Parameters $\{w_{c'ci}^{(\ell)}\}$ and parameters in $f_{\mathrm{U}}$, $f_{\mathrm{V}}$
are common for all matrices.
Prior mean values $\vec{U}^{(0)}$, $\vec{V}^{(0)}$ are different across matrices
since they are calculated from input matrices $\vec{X}$ and $\vec{B}$. 

We estimate the factorized matrices
by the MAP estimation in Eq.~(\ref{eq:map}) using the prior mean values in Eq.~(\ref{eq:uv0}) as
the initial values.
The update rules of the MAP estimation based on the gradient descent method
are given in a closed form by taking the gradient of the objective function
with respect to factorized matrices
$\vec{u}_{n}$ and $\vec{v}_{m}$:
\begin{align}
  \vec{u}^{(t+1)}_{n}=\vec{u}^{(t)}_{n}-\eta\Bigl(\sum_{m=1}^{M}b_{nm}(\vec{u}_{n}^{(t)\top}
  \vec{v}^{(t)}_{m}-x_{nm})\vec{v}^{(t)}_{m}
  +\lambda(\vec{u}^{(t)}_{n}-\vec{u}^{(0)}_{n})\Bigr),
  \label{eq:u}
\end{align}
\begin{align}
  \vec{v}^{(t+1)}_{m}=\vec{v}^{(t)}_{m}-\eta\Bigl(\sum_{n=1}^{N}b_{nm}(\vec{u}_{n}^{(t)\top}
  \vec{v}^{(t)}_{m}-x_{nm})\vec{u}^{(t)}_{n}
  +\lambda(\vec{v}^{(t)}_{m}-\vec{v}^{(0)}_{m})\Bigr),
  \label{eq:v}  
\end{align}
where $\vec{u}^{(t)}$ and $\vec{v}^{(t)}$ are estimated
factorized matrices at the $t$th iteration,
and $\eta>0$ is the learning rate.
The factorized matrices at the $T$th iteration
are used as the output of our model
$\vec{U}(\vec{X},\vec{B})=\vec{U}^{(T)}$, $\vec{V}(\vec{X},\vec{B})=\vec{V}^{(T)}$.
The missing value in $\vec{X}$ is predicted by the inner product of the factorized matrices
by
\begin{align}
  \hat{x}_{nm}=\vec{u}_{n}(\vec{X},\vec{B})^{\top}\vec{v}_{m}(\vec{X},\vec{B}).
\end{align}

We could predict the missing values using
the output of the neural networks,
$\hat{x}_{nm}=\vec{u}_{n}^{(0)\top}\vec{v}_{m}^{(0)}$,
without the gradient descent steps.
However, the prediction might be different from the true values
even with the observed elements
if the given matrix does not resemble any of the meta-training matrices
that are used for optimizing the neural networks,
and the generalization performance of the neural networks
is not high enough.
With gradient descent steps,
the factorized matrices can be adapted to the given matrix
even when the neural networks fail to adapt.
Minimizing the error between the observed and predicted values
is a standard technique for matrix factorization~\cite{koren2009matrix}.
We use it as a component in our model.

Our model can be seen as a single neural network,
that takes a matrix with missing values as input,
and outputs its factorized matrices,
where the exchangeable matrix layers in Eq.~(\ref{eq:z}),
the permutation invariant networks in Eq.~(\ref{eq:uv0}),
and the gradient descent steps in Eqs.~(\ref{eq:u},\ref{eq:v})
are used as layers (Figure~\ref{fig:model}).
Algorithm~\ref{alg:model} shows the forwarding procedures of our model.
Since our model including the gradient descent steps is differentiable,
we can backpropagate the loss through the gradient descent steps
to update the parameters in our neural networks.
  
\begin{algorithm}[t!]
  \caption{Forwarding procedures of our model.}
  \label{alg:model}
  \begin{algorithmic}[1]
    \renewcommand{\algorithmicrequire}{\textbf{Input:}}
    \renewcommand{\algorithmicensure}{\textbf{Output:}}
    \REQUIRE{Observed matrix with missing values $\vec{X}$, indicator matrix $\vec{B}$, number of iterations $T$}
    \ENSURE{Factorized matrices $\vec{U}(\vec{X},\vec{B})$, $\vec{V}(\vec{X},\vec{B})$}
    \STATE Obtain element representations $\vec{Z}$ by Eq.~(\ref{eq:z}).
    \STATE Calculate the mean of the priors of factorized matrices $\vec{U}^{(0)}$ and $\vec{V}^{(0)}$ by Eq.~(\ref{eq:uv0}).
    \FOR{$t:=1$ to $T$}
    \STATE Update factorized matrices $\vec{U}^{(t)}$, $\vec{V}^{(t)}$ by Eqs.~(\ref{eq:u},\ref{eq:v}) based on the MAP estimation.
    \ENDFOR
  \end{algorithmic}
\end{algorithm}

\subsection{Meta-training}
\label{sec:training}

We meta-train model parameters $\bm{\Theta}$,
i.e., exchangeable matrix layer parameters $\{\{\{\{\{w^{(\ell)}_{cc'i}\}_{i=1}^{5}\}_{c'=1}^{C^{\ell}}\}_{c=1}^{C^{\ell+1}}\}_{\ell=1}^{L-1}$,
the parameters of feed-forward neural networks $f_{\mathrm{U}}$ and $f_{\mathrm{V}}$,
and regularization parameter $\lambda$,
by minimizing the following expected test error
of the missing values with the episodic training framework:
\begin{align}
  \mathbb{E}_{\vec{X},\vec{B},\vec{X}',\vec{B}'}[\parallel
    \vec{B}'\odot(\vec{X}'-\vec{U}(\vec{X},\vec{B})^{\top}\vec{V}(\vec{X},\vec{B}))\parallel^{2}],
  \label{eq:error}
\end{align}
where $\vec{X}$ and $\vec{X}'$ are sampled
training and test matrices,
$\vec{B}$ and $\vec{B}'$ are their indicator matrices,
$\vec{U}(\vec{X},\vec{B})\in\mathbb{R}^{N\times K}$ and
$\vec{V}(\vec{X},\vec{B})\in\mathbb{R}^{M\times K}$ are
the factorized matrices obtained by our model
from training matrix $\vec{X}$ and its indicator matrix $\vec{B}$,
$\mathbb{E}$ is the expectation,
and $\odot$ is the element-wise multiplication.
Eq.~(\ref{eq:error})
is the expectation error between the test matrix 
and its imputation by our model adapted to the training matrix.

Algorithm~\ref{alg:train} shows the meta-training procedures of our model.
In Line 4,
matrix $\bar{\vec{X}}\in\mathbb{R}^{N\times M}$
is constructed from randomly selected meta-training matrix $\vec{X}_{d}$,
where $\bar{\vec{X}}$ is a submatrix of $\vec{X}_{d}$.
Instead of sampling the submatrices,
we can use the whole selected meta-training matrix $\bar{\vec{X}}=\vec{X}_{d}$,
or change the number of rows and columns,
$N$ and $M$, for each epoch.
In Line 5, the non-missing elements in matrix $\bar{\vec{X}}$
are randomly split into training matrix $\vec{X}$
and test matrix $\vec{X}'$,
where their indicator matrices are mutually exclusive,
$\vec{B}\cap\vec{B}'=\phi$.
We assume that the meta-test matrices are missing at random.
If they are missing not at random,
we can model missing patterns~\cite{marlin2007collaborative},
and use them for generating training and test matrices in the meta-training procedures.

The time complexity for each meta-training step is $O(\gamma LNM+T(\gamma NM+(N+M)K)+\gamma'NM)$,
where $\gamma$ is the rate of the observed elements in a training matrix,
and $\gamma'$ is that in a test matrix.
The first term $O(\gamma LNM)$ is for inferring the priors by the neural networks.
The second term $O(T(\gamma NM+(N+M)K))$ is for the MAP estimation of
the factorized matrices using Eqs.~(\ref{eq:u},\ref{eq:v}).
The third term $O(\gamma'NM)$ is for calculating the loss on the test matrix.
The number of model parameters $\bm{\Theta}$ depends on neither the meta-training data size
nor the numbers of rows $N$ and columns $M$ of the training and test matrices.

\begin{algorithm}[t!]
  \caption{Meta-training procedure of our model.}
  \label{alg:train}
  \begin{algorithmic}[1]
    \renewcommand{\algorithmicrequire}{\textbf{Input:}}
    \renewcommand{\algorithmicensure}{\textbf{Output:}}
    \REQUIRE{Meta-training data $\{\vec{X}_{d}\}_{d=1}^{D}$,
      number of rows $N$, number of columns $M$,
      training ratio $R$, number of iterations $T$}
    \ENSURE{Trained model parameters $\bm{\Theta}$}
    \STATE Initialize model parameters $\bm{\Theta}$.
    \WHILE{End condition is satisfied}
    \STATE Randomly select task index $d$ from $\{1,\cdots,D\}$.
    \STATE Randomly sample $N$ rows and $M$ columns from $\vec{X}_{d}$ and construct matrix $\bar{\vec{X}}$.
    \STATE Randomly assign non-missing elements
    in $\bar{\vec{X}}$ with probability $R$
    as training matrix $\vec{X}$, and assign the
    remaining non-missing elements as test matrix $\vec{X}'$.
    \STATE Obtain factorized matrices $\vec{U}(\vec{X},\vec{B}), \vec{V}(\vec{X},\vec{B})$ of the training matrix by Algorithm~\ref{alg:model}.
    \STATE Calculate loss $\parallel\vec{B}'\odot(\vec{X}'-\vec{U}(\vec{X},\vec{B})^{\top}\vec{V}(\vec{X},\vec{B}))\parallel^{2}$ on the test matrix and its gradient.
    \STATE Update model parameters $\bm{\Theta}$ using the loss and gradient by a stochastic gradient method.
    \ENDWHILE
  \end{algorithmic}
\end{algorithm}

\section{Experiments}
\label{sec:experiments}

\subsection{Data}

We evaluated the proposed method using three rating datasets:
ML100K, ML1M~\cite{harper2015movielens}, and Jester~\cite{goldberg2001eigentaste}~\footnote{ML100K and ML1M were obtained from \url{https://grouplens.org/datasets/movielens/},
  and Jester was obtained from \url{https://goldberg.berkeley.edu/jester-data/}.}.
The ML100K data contained 100,000 ratings of 1,682 movies by 943 users.
The ML1M data contained 1,000,209 ratings of 3,952 movies by 6,040 users.
The Jester data contained 1,805,072 ratings of 100 jokes from 24,983 users.
The ratings of each dataset were normalized with zero mean and unit variance.
We randomly split the users and items,
and used 70\% of them for meta-training,
10\% for meta-validation,
and the remaining for meta-test.
There were no overlaps of users and items across the meta-training,
validation, and test data.
We randomly generated ten $30 \times 30$ meta-test matrices from the meta-test data
and used half of the originally observed ratings as
missing ratings for evaluation.
For each meta-test matrix, the average numbers of observed elements were respectively 28.8, 19.0, and 218.0
in the ML100K, ML1M, and Jester data.
We did not use the meta-test matrices in the meta-training phase.
The evaluation measurement was the test mean squared error,
which was calculated by the mean squared error between the true and estimated ratings
of the unobserved elements in the meta-test matrices.
We averaged the test mean squared errors
over ten experiments with different splits of meta-training, validation, and test data.

\subsection{Proposed method setting}

We used three exchangeable matrix layers with 32 channels.
Feed-forward neural networks $f_{\mathrm{u}},f_{\mathrm{v}}$
were four-layered with 32 hidden units and $K=32$ output units.
We used a rectified linear unit, $\mathrm{ReLU}(x)=\max(0,x)$, for the activation.
For the gradient descent steps of the matrix factorization in Eqs.~(\ref{eq:u},\ref{eq:v}),
the learning rate was $\eta=10^{-2}$,
and the number of iterations was $T=10$.
The number of rows and columns of the training matrices were $N=30$ and $M=30$,
and the training ratio was $R=0.5$.
We optimized our model using Adam~\cite{kingma2014adam} with learning rate $10^{-4}$,
batch size $16$, and dropout rate $0.1$~\cite{srivastava2014dropout}.
The number of meta-training epochs was 1,000, and
the meta-validation data were used for early stopping.
We implemented the proposed method with PyTorch~\cite{paszke2017automatic}.

\subsection{Comparing methods}

We compared the proposed method with the following eight methods:
exchangeable matrix layer neural networks (EML)~\cite{hartford2018deep},
EML finetuned with the meta-test matrix (FT),
model-agnostic meta-learning of EML (MAML)~\cite{finn2017model},
item-based AutoRec (AutoRecI)~\cite{sedhain2015autorec},
user-based AutoRec (AutoRecU),
deep matrix factorization (DMF)~\cite{xue2017deep},
matrix factorization (MF),
and the mean value of the meta-test matrix (Mean).
EML, FT, MAML as well as the proposed method
were meta-learning schemes,
all of which were trained using the meta-training matrices
such that the test mean squared error was minimized.
AutoRecI, AutoRecU, DMF, MF, and Mean
were trained using the observed ratings of the meta-test matrix
by minimizing the mean squared error
without meta-training matrices.

EML used three exchangeable matrix layers,
where the number of channels with the first two layers was 32,
and the number of channels with the last layer was one to output the estimation of a rating.
EML was trained with the episodic framework like the proposed method.
The exchangeable matrix layers have not been used for meta-learning.
We newly used them for meta-learning 
by employing them as the encoder and decoder
in an encoder-decoder based meta-learning method.
FT finetuned the parameters of the trained EML
using the observed ratings in the meta-test matrix
by minimizing the mean squared error.
MAML trained the initial parameters of EML
to minimize the test mean squared error when finetuned
by the episodic training framework.
The number of iterations in the inner loop was five.
AutoRec is a neural network-based matrix imputation method.
With AutoRecI (AutoRecU),
a neural network took each row (column) as input
and output its reconstruction.
DMF is a neural network-based matrix factorization method,
where the row and column representations were calculated by neural networks
that took a row or column as input,
and ratings were estimated by the inner product of the row and column representations.
For the neural networks in AutoRecI, AutoRecU, and DMF,
we used four-layered feed-forward neural networks with 32 hidden units.
With DMF and MF, the number of latent factors was 32.
With AutoRecI, AutoRecU, DMF, and MF,
the weight decay parameter was tuned from $\{10^{-4},10^{-3},10^{-2},10^{-1},1\}$,
and the learning rate was tuned from $\{10^{-3},10^{-2},10^{-1}\}$
using the validation data.

\subsection{Results}

\begin{table}[t!]
  \centering  
  \caption{Average test mean squared errors on unobserved elements in test matrices and their standard error: Values in bold are not statistically different at 5\% level from the best performing method in each dataset by a paired t-test.}
  \label{tab:mse}
  \begin{tabular}{lrrr}
    \hline    
    & ML100K & ML1M & Jester\\
    \hline
    Ours & {\bf 0.901 $\pm$ 0.033} &{\bf 0.883 $\pm$ 0.024} &{\bf 0.813 $\pm$ 0.009} \\
    EML & 0.933 $\pm$ 0.036 & 0.907 $\pm$ 0.024 & 0.848 $\pm$ 0.009 \\
    FT & 1.175 $\pm$ 0.047 & 1.149 $\pm$ 0.046 & 0.990 $\pm$ 0.008 \\
    MAML & 0.941 $\pm$ 0.036 & 0.904 $\pm$ 0.025 & 0.880 $\pm$ 0.011 \\
    AutoRecI & 0.985 $\pm$ 0.040 & 0.968 $\pm$ 0.028 & 0.949 $\pm$ 0.008 \\
    AutoRecU & 0.987 $\pm$ 0.033 & 0.962 $\pm$ 0.024 & 0.907 $\pm$ 0.014 \\    
    DMF & 0.979 $\pm$ 0.034 & 0.972 $\pm$ 0.023 & 0.852 $\pm$ 0.007 \\
    MF & 1.014 $\pm$ 0.037 & 0.962 $\pm$ 0.031 & 1.005 $\pm$ 0.014 \\
    Mean & 1.007 $\pm$ 0.020 & 0.983 $\pm$ 0.013 & 1.004 $\pm$ 0.008\\
 \hline
\end{tabular}
\end{table}

\begin{figure*}[t!]
  \centering
  {\tabcolsep=0.1em
  \begin{tabular}{ccc}
    \includegraphics[width=13em]{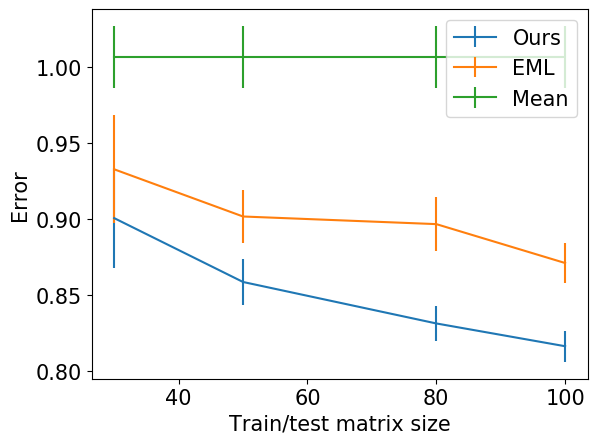} &
    \includegraphics[width=13em]{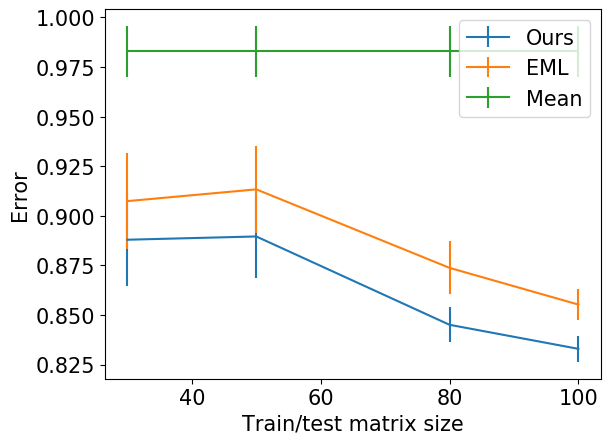} &
    \includegraphics[width=13em]{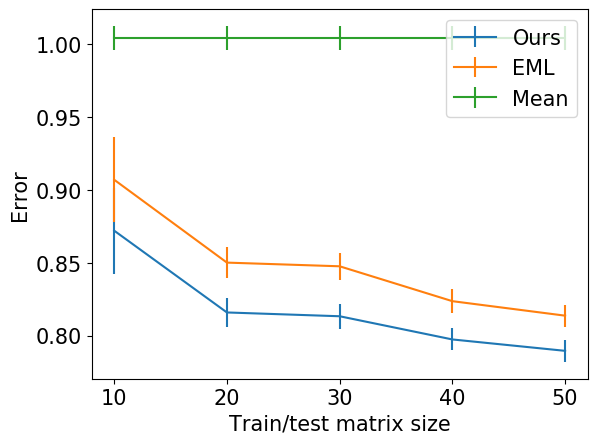} \\
    (a) ML100K & (b) ML1M & (c) Jester\\
  \end{tabular}}
  \caption{Average test mean squared errors
    when meta-trained with training and test matrices of different sizes,
    where the size of the meta-test matrices was the same as that of the training and test matrices.
    We used square matrices, and the horizontal axis is the number of columns
  and rows.
  Bar show the standard error.}
  \label{fig:error_matrix_size}
\end{figure*}

\begin{figure*}[t!]
  \centering
  {\tabcolsep=0.1em  
  \begin{tabular}{ccc}
    \includegraphics[width=13em]{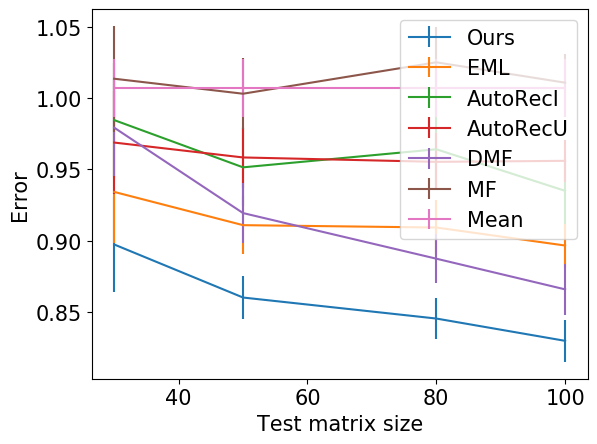} &
    \includegraphics[width=13em]{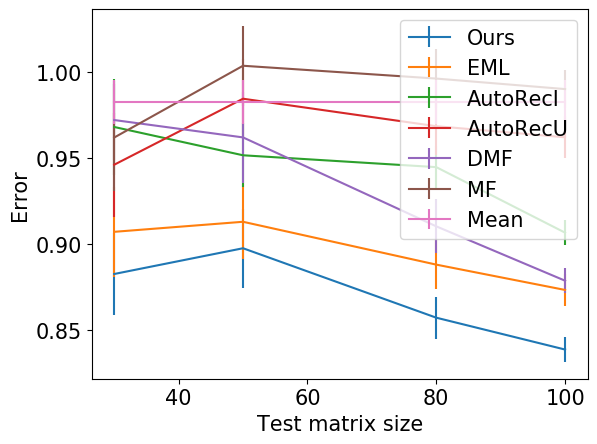} &
    \includegraphics[width=13em]{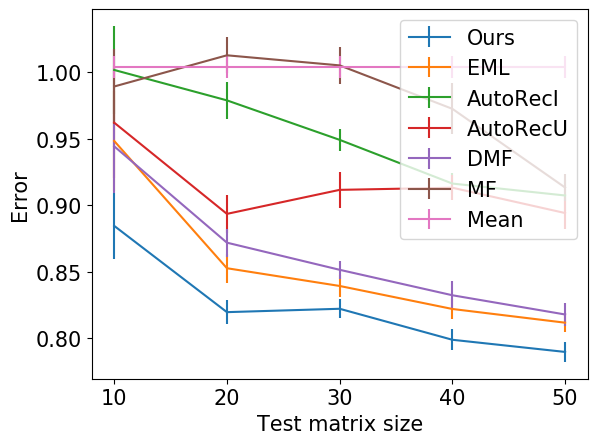} \\
    (a) ML100K & (b) ML1M & (c) Jester\\
  \end{tabular}}
  \caption{Average test mean squared errors
    with meta-test matrices with different sizes,
    where the model was meta-trained with $30 \times 30$ matrices.
  We used square matrices, and the horizontal axis is the number of columns
  and rows.    
    Bars show the standard error.}
  \label{fig:error_matrix_size_test}
\end{figure*}

Table~\ref{tab:mse} shows the average test mean squared error.
The proposed method achieved the lowest error on all the datasets.
EML's performance was the second best on the ML100K and Jester datasets.
This result indicates that exchangeable matrix layers
effectively extracted useful information from the matrices
with missing values.
The proposed method further improved the performance from EML
by directly adapting to the observed elements
using the gradient descent steps based on MAP estimation.
Since EML approximates the adaptation to the observed elements
only by exchangeable matrix layers,
the estimated values can be different from the observed elements.
The errors on the observed elements with EML were higher
than those with the proposed method shown in the supplementary material.
With FT, although the errors on the observed elements were low,
the test errors on the unobserved elements were high,
because FT was overfitted to the observed values by
training the model by minimizing the error on the observed elements.
On the other hand, the proposed method
trains the model by minimizing the test error on the unobserved elements
when fitted to the observed elements with MAP estimation.
Therefore, the proposed method alleviated overfitting
to the observed elements.
Since MAML trained the model by minimizing the expected test error,
the overfitting was smaller than FT.
However, MAML's performance did not surpass that of EML
and was lower than that of the proposed method.
With MAML, the whole neural network-based model is adapted
to the observed values.
In contrast, the proposed method adapts only factorized matrices
to the observed values,
although the neural networks are fixed and used for defining the priors
of the factorized matrices.
The errors of the methods that did not use meta-training matrices, i.e.,
AutoRec, DMF, MF, and Mean, were high
since they needed to estimate the missing values only from 
a small number of observations.
On the other hand,
the proposed method achieved the lowest error
by meta-learning hidden structure in the meta-training matrices
that effectively improved the test matrix imputation performance
even though rows and columns were not shared across different matrices.

Figure~\ref{fig:error_matrix_size}
shows the test mean squared errors when
meta-trained with training and test matrices of different sizes,
where the size of the meta-test matrices was the same as that of the training and test matrices.
Our proposed method and EML decreased the error
as the matrix size increased
because the number of observations rose as the matrix size increased.
Figure~\ref{fig:error_matrix_size_test}
shows the test mean squared errors
with meta-test matrices of different sizes,
where the model was meta-trained with $30 \times 30$ matrices.
The proposed method achieved the lowest error with different sizes
of meta-test matrices.
The proposed method's performance improved
as the meta-test matrix size increased even though it
was trained with different-sized matrices.

\begin{figure*}[t!]
  \centering
  {\tabcolsep=0.1em  
  \begin{tabular}{ccc}
    \includegraphics[width=12em]{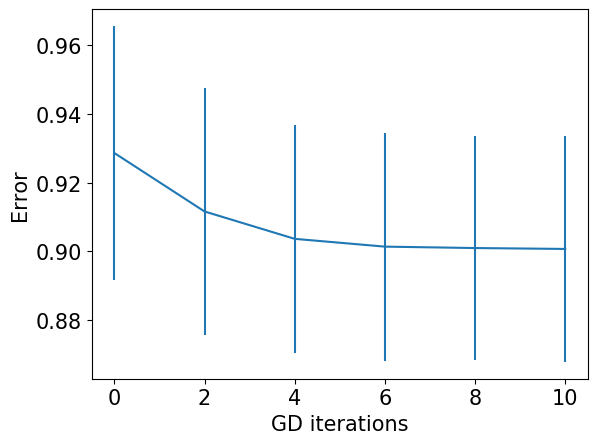} &
    \includegraphics[width=12em]{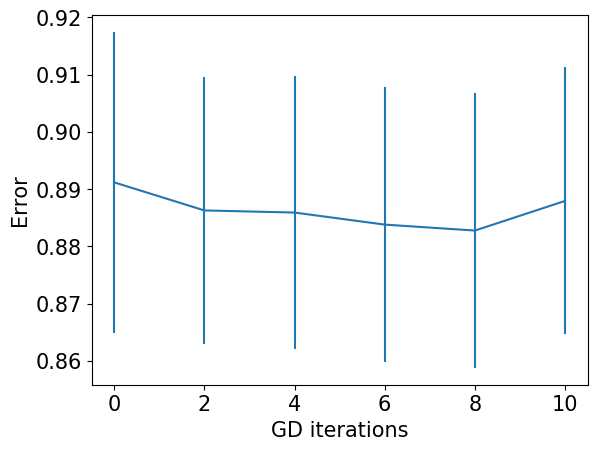} &
    \includegraphics[width=12em]{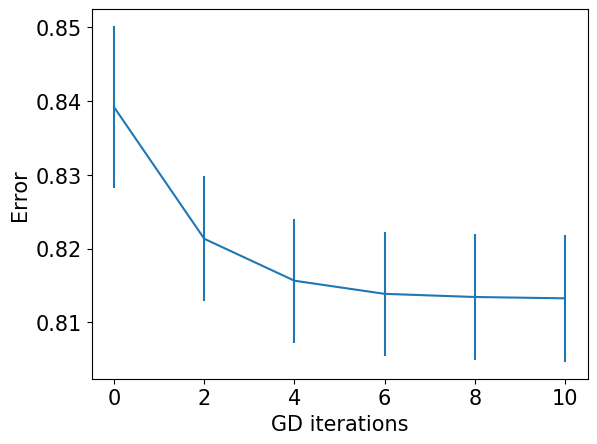} \\
    (a) ML100K & (b) ML1M & (c) Jester\\
  \end{tabular}}
  \caption{Average test mean squared errors
    with different numbers of gradient descent iterations with the proposed method. Bar show the standard error.}
  \label{fig:error_gd}
\end{figure*}

Figure~\ref{fig:error_gd} shows the test mean squared errors with different numbers of gradient descent iterations with the proposed method. As the number of iterations increased, the error decreased especially with the ML100K and Jester data. This result indicates the effectiveness of the gradient descent steps in the proposed method.

Table~\ref{tab:train_mse} shows the average training
mean squared errors.
The errors on the observed elements with EML were higher
than those with the proposed method.

Table~\ref{tab:mse_ablation} shows
the average test mean squared errors by the proposed method
when different datasets were used between meta-training and meta-test data.
With the ML1M and Jester meta-test data,
the proposed method achieved the best performance
when meta-trained with matrices from the same dataset.
Even when the meta-training datasets were different from the meta-test datasets,
the errors were relatively low, and
they were lower than those by the comparing methods.
This is because the datasets used in our experiments were
related to each other and shared some hidden structure, and 
the proposed method learned the shared structure and used the learned structure
to improve the matrix imputation performance in the other datasets.

Table~\ref{tab:mse_K} shows
the average test mean squared errors by the proposed method
with different factorized matrix ranks $K$.
The proposed method worked even when factorized matrix rank $K$ is larger
than the number of rows $N$ or columns $M$
since it uses the priors inferred by the neural networks based on the MAP estimation.
Factorized matrix rank $K$ slightly affects the performance,
but the proposed method achieved better performance than the comparing methods
with a wide range of $K$.

Table~\ref{tab:mse_autorec} shows
the average test mean squared errors by AutoRec
with different numbers of hidden units.
With any numbers of hidden units,
the performance by AutoRec was worse than the proposed method.

Tables~\ref{tab:train_time} and \ref{tab:test_time}
show the computational time in seconds for meta-training and test
on computers with 2.60GHz CPUs.
The meta-training time of the proposed method was much shorter than MAML
since the proposed method adapts only factorized matrices
instead of neural networks,
where the adaption steps can be written explicitly.
The meta-training time of the proposed method was longer than EML
since the proposed method additionally requires adaptation steps.
The proposed method's meta-test time was short since
it requires only a small number of adaptation steps for a few observed elements.

\begin{table}[h]
  \centering  
  \caption{Average mean squared errors on observed elements in test matrices and their standard error with meta-learning methods.}
  \label{tab:train_mse}
  \begin{tabular}{lrrr}
    \hline    
    & ML100K & ML1M & Jester\\
    \hline
Ours & 0.504 $\pm$ 0.017 &0.354 $\pm$ 0.013 &0.618 $\pm$ 0.008 \\
EML & 0.596 $\pm$ 0.018 & 1.573 $\pm$ 0.071 & 0.664 $\pm$ 0.010 \\
FT & 0.328 $\pm$ 0.013 & 0.346 $\pm$ 0.017 & 0.528 $\pm$ 0.012 \\
MAML & 0.493 $\pm$ 0.012 & 0.503 $\pm$ 0.028 & 0.658 $\pm$ 0.013 \\
 \hline
\end{tabular}
\end{table}

\begin{table}[t!]
  \centering  
  \caption{Average test mean squared errors by the proposed method with different pairs of meta-training and meta-test datasets.
    Each row represents the meta-training data, and each column represents the meta-test data.}
  \label{tab:mse_ablation}
  \begin{tabular}{lrrr}
    \hline    
    Meta-training data $\backslash$ Meta-test data & ML100K & ML1M & Jester\\
    \hline
    ML100K & 0.901 $\pm$ 0.033 & 0.906 $\pm$ 0.023 & 0.825 $\pm$ 0.008 \\
    ML1M & 0.889 $\pm$ 0.028 & 0.883 $\pm$ 0.024 & 0.827 $\pm$ 0.008 \\
    Jester & 0.900 $\pm$ 0.027 & 0.927 $\pm$ 0.025 & 0.813 $\pm$ 0.009 \\
    ML100K, ML1M & 0.894 $\pm$ 0.036 & 0.883 $\pm$ 0.024 & 0.829 $\pm$ 0.008 \\
    ML100K, Jester & 0.894 $\pm$ 0.031 & 0.893 $\pm$ 0.025 & 0.824 $\pm$ 0.008 \\
    ML1M, Jester & 0.884 $\pm$ 0.035 & 0.885 $\pm$ 0.024 & 0.832 $\pm$ 0.008 \\
    ML100K, ML1M, Jester & 0.892 $\pm$ 0.035 & 0.884 $\pm$ 0.023 & 0.829 $\pm$ 0.008\\
    \hline
  \end{tabular}
\end{table}

\begin{table}[t!]
  \centering  
\caption{Average test mean squared errors by the proposed method with different factorized matrix ranks $K$.}
  \label{tab:mse_K}
\begin{tabular}{lrrrrr}
  \hline
$K$ & 8 & 16 & 32 & 64 & 128\\
  \hline
  ML100K & 0.908 & 0.903 & 0.901 & 0.899 & 0.898 \\
  ML1M & 0.889 & 0.887 & 0.883 & 0.888 & 0.886 \\
  Jester & 0.815 & 0.814 & 0.813 & 0.814 & 0.813 \\
  \hline
\end{tabular}
\end{table}

\begin{table}[t!]
  \centering  
\caption{Average test mean squared errors by AutoRec with different numbers of hidden units.}
  \label{tab:mse_autorec}
\begin{tabular}{lrrrr}
  \hline
  & \multicolumn{2}{c}{AutoRecI} & \multicolumn{2}{c}{AutoRecU} \\
  \hline
  \#hidden units & 128 & 512 & 128 & 512 \\
  \hline
  ML100K & 0.980 & 0.974 & 0.998 & 0.979 \\
  ML1M & 0.949 & 0.954 & 0.946 & 0.947 \\
  Jester & 0.924 & 0.929 & 0.876 & 0.917 \\
  \hline
\end{tabular}
\end{table}

\begin{table}[t!]
  \centering  
  \caption{Meta-training time in seconds.}
  \label{tab:train_time}
  \begin{tabular}{lrrr}
    \hline    
    & ML100K & ML1M & Jester\\
    \hline
Ours & 1039 & 3527 & 376 \\
EML & 850 & 2983 & 247 \\
MAML & 15983 & 55205 & 4189 \\
\hline
\end{tabular}
\end{table}

\begin{table}[t!]
  \centering  
  \caption{Meta-test time in seconds.}
  \label{tab:test_time}
  \begin{tabular}{lrrr}
    \hline    
    & ML100K & ML1M & Jester\\
    \hline
Ours & 0.13 & 0.13 & 0.14 \\
EML & 0.09 & 0.09 & 0.10 \\
FT & 3.88 & 3.75 & 2.77 \\
MAML & 1.72 & 3.12 & 1.60 \\
AutoRecI & 1.51 & 1.54 & 3.60 \\
AutoRecU & 1.50 & 1.49 & 3.99 \\
DMF & 1.78 & 5.20 & 4.49 \\
MF & 0.53 & 1.20 & 1.29\\
\hline
\end{tabular}
\end{table}

\section{Conclusion}
\label{sec:conclusion}

We proposed a neural network-based meta-learning method
for matrix imputation that
learns from multiple matrices without shared rows and columns,
and predicts the missing values given observations in unseen matrices.
Although we believe that our work is an important step
for learning from a wide variety of matrices,
we must extend our approach in several directions.
First, we will apply our framework to tensor data
using exchangeable tensor layers~\cite{hartford2018deep}
and tensor factorizations~\cite{hitchcock1927expression,carroll1970analysis,harshman1970foundations,tucker1966some,welling2001positive,kuleshov2015tensor}.
Second, we will use our framework for other types of matrix factorization,
such as non-negative matrix factorization (NMF)~\cite{lee1999learning}
and independent component analysis~\cite{hyvarinen2000independent}.
For example,
we can use multiplicative update steps for NMF
instead of gradient descent steps.
Third, we want to extend our proposed method to use auxiliary information,
e.g., user and item information in recommender systems,
by taking it as input of our neural network.

\bibliographystyle{abbrv}
\bibliography{arxiv_meta_mf}

\end{document}